\documentclass[sigconf]{acmart}
\makeatletter
\renewcommand\@formatdoi[1]{\ignorespaces}
\makeatother

\AtBeginDocument{%
  \providecommand\BibTeX{{%
    \normalfont B\kern-0.5em{\scshape i\kern-0.25em b}\kern-0.8em\TeX}}}

\setcopyright{rightsretained}
\copyrightyear{2023}
\acmYear{2023}
\acmDOI{XXXXXXX.XXXXXXX}

\acmConference[SIGIR KDF’23]{The SIGIR'23 Workshop on Knowledge Discovery from Unstructured Data in Financial Services}{July 27,
  2023}{Taipei, Taiwan}

\acmPrice{15.00}
\acmISBN{}

\usepackage{fontawesome5}
\usepackage{hyperref}

\usepackage{xcolor}
\usepackage[normalem]{ulem}
\useunder{\uline}{\ulined}{}%
\DeclareUrlCommand{\bulurl}{}

\begin{document}

\title{FinTree \faTree: Financial Dataset Pretrain Transformer Encoder for Relation Extraction}

\author{Hyunjong Ok}
\affiliation{
  \institution{Kyung Hee University}
  \city{Yongin}
  \country{Korea}}
\email{r7play@khu.ac.kr}

\renewcommand{\shortauthors}{Hyunjong Ok.}

\begin{abstract}
We present FinTree, \textbf{Fin}ancial Dataset Pretrain \textbf{T}ransformer Encoder for
\textbf{Re}lation \textbf{E}xtraction. Utilizing an encoder language model, we further pretrain FinTree on the financial dataset, adapting the model in financial domain tasks. FinTree stands out with its novel structure that predicts a masked token instead of the conventional [CLS] token, inspired by the Pattern Exploiting Training methodology. This structure allows for more accurate relation predictions between two given entities. The model is trained with a unique input pattern to provide contextual and positional information about the entities of interest, and a post-processing step ensures accurate predictions in line with the entity types. Our experiments demonstrate that FinTree outperforms on the REFinD, a large-scale financial relation extraction dataset. 
The code and pretrained models are available at \bulurl{https://github.com/HJ-Ok/FinTree}.
\end{abstract}



\maketitle

\section{Introduction}
The field of finance has witnessed an explosion of text data, from news articles and social media posts to financial reports and analyst notes. This wealth of data has opened new research opportunities, especially in NLP. Relation Extraction (RE), a task to identify and classify semantic relationships between entities in text, is a primary problem in NLP. The importance of this task in the financial domain is also grown up, as it can get valuable insights from the unstructured financial domain texts. However, despite the development of NLP techniques, relation extraction in the financial domain poses unique challenges. Financial text inherently differs from general text's semantics, terminologies, and writing style. So numerous state-of-the-art (SOTA) relation extraction models encounter significant challenges when applied to relation extraction of the financial domain texts \cite{kaur2023refind}. With our study, we aim to address this problem, providing a novel approach to relation extraction in financial texts and making a clear contribution to this area.

Our work presents the task-specific model, the `FinTree: \textbf{Fin}ancial Dataset Pretrain \textbf{T}ransformer Encoder for \textbf{Re}lation \textbf{E}xtraction.' We leverage the Transformer Encoder model \cite{NIPS2017_3f5ee243} performance in financial domain tasks through pretraining on a financial dataset. In contrast to conventional classification tasks, we adopt a slightly divergent approach. We orient our model's prediction towards the [MASK] token rather than the [CLS] token. This technique, similar to the Pattern Exploiting Training (PET) \cite{schick-schutze-2021-exploiting}, aligns the task with the masked language modeling (MLM) pretraining methods. By focusing on predicting masked tokens, our approach facilitates smoother transfer learning during finetuning, boosting performance. We guide our model by providing a specially formatted text input that encourages focus on relevant entities and their relations. We also post-process to ensure the model accurately predicts relations involving the correct entities.

The proposed approach of FinTree has a strong performance in the financial domain relation extraction, provides insights into the challenges of financial domain relation extraction, and contributes to further research in this area. 

\begin{figure}[t]
    \centering
    \includegraphics[width=1.0\columnwidth]{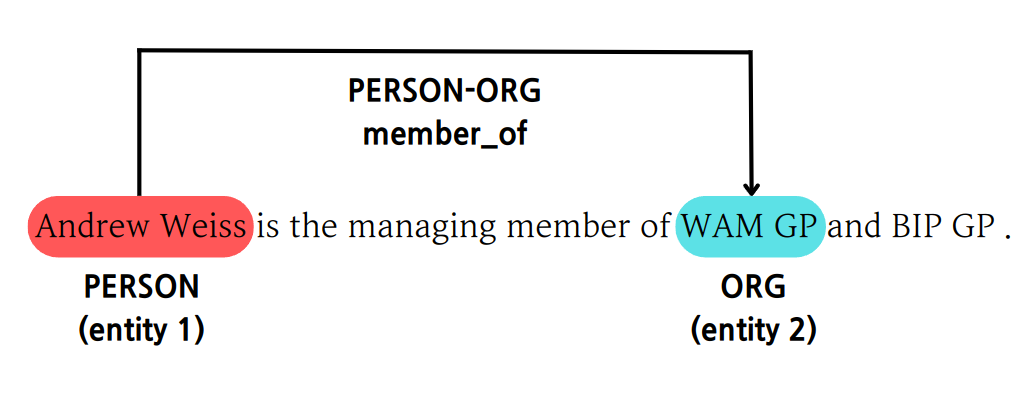}
    \caption{Examples of Relation Extraction in the Financial Domain. Two entities, along with their types, are provided. The task is to predict the relationship between them.}
    \label{fig:overview}
\end{figure}

\section{Methods}
\subsection{Model Selection}
\begin{figure*}[htp]
    \centering
    \includegraphics[width=17.5cm]{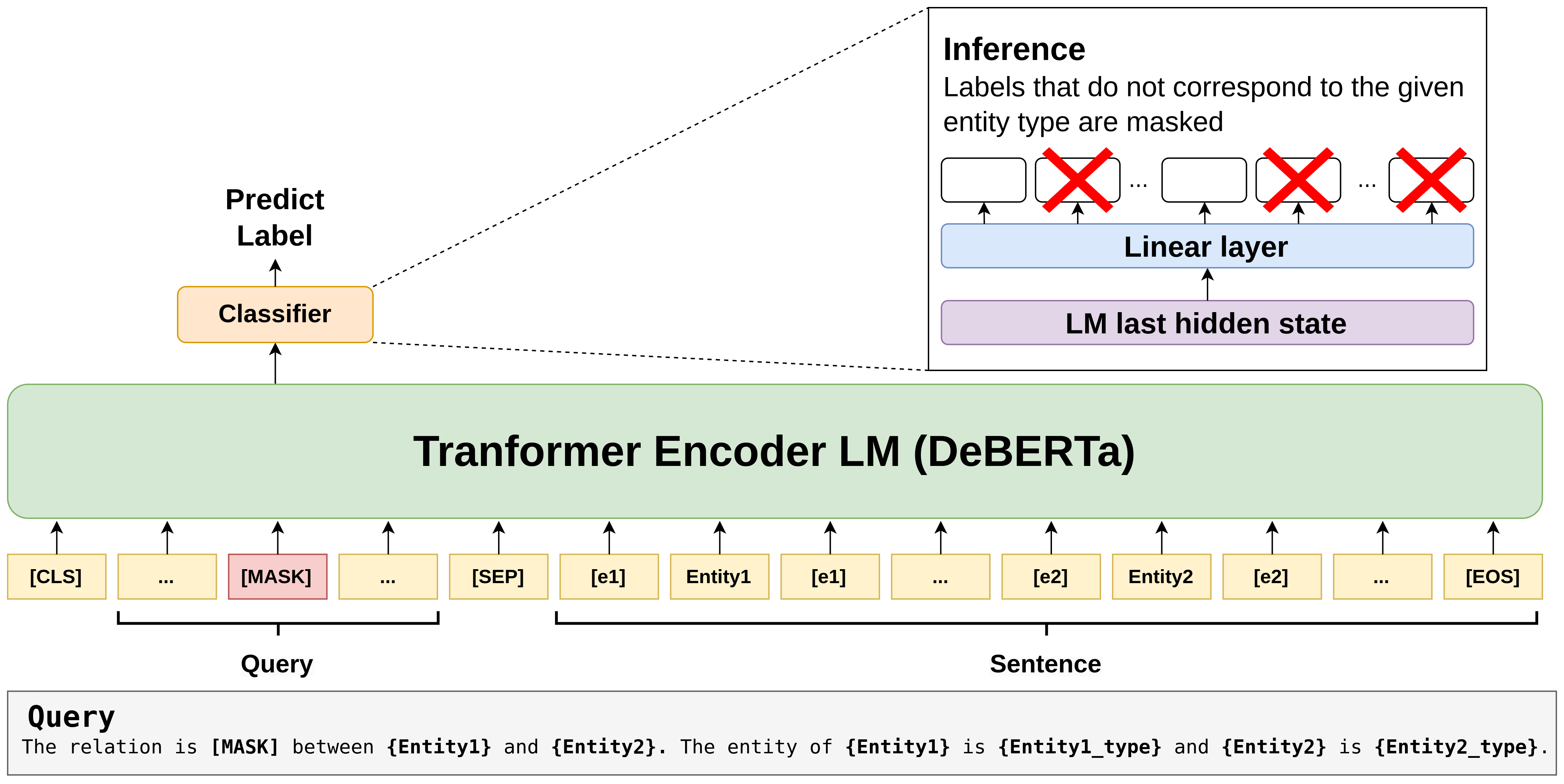}
    \caption{The overall structure of our Relation Extraction model: FinTree}
    \label{fig_1}
\end{figure*}
In designing our approach for relation extraction in financial documents, we first conducted an exploratory analysis to identify an appropriate pretrained language model to serve as our foundational backbone. The models evaluated in this study included well-known architectures in the NLP field, known for their exemplary performance in various tasks. Specifically, we considered BERT \cite{devlin-etal-2019-bert}, RoBERTa \cite{liu2019roberta}, ALBERT \cite{Lan2020ALBERT:}, and DeBERTa \cite{he2022debertav3} in their Base and Large configurations. All chosen language models were trained on the same REFinD train dataset with the same hyperparameters to get fairness. We measure the performance using the development dataset. We adopted F1-score metrics in macro, micro, and weighted configurations to evaluate the models comprehensively.

Our exploratory analysis revealed that DeBERTa consistently outperformed the other models. A detailed exposition of the performance metrics for each model is in Table~\ref{tab:model selection result}. Accordingly, we used DeBERTa as our language model backbone for all subsequent experiments.

\subsection{Further Pretraining}
In our methodological approach, we enhance the capabilities of the Transformer Encoder model by pretraining it on a financial corpus, making it more adaptable to the financial domain's unique semantics, terminologies, and writing style.

We use MLM pretraining methods similar to the method in BERT. We collect a large corpus of financial texts for additional pretraining, the EDGAR-CORPUS \cite{loukas-etal-2021-edgar} and U.S. Securities and Exchange Commission (SEC) Financial Statement and Notes Data Sets\footnote{https://www.sec.gov/about/divisions-offices/division-economic-risk-analysis/data/financial-statement-and-notes-data-set}. We employed a pretraining dataset with text lengths ranging from 64 to 2048. This selection ensures that the pretraining data exhibit a similar distribution of sequence lengths to the target dataset, REfinD. By aligning the sequence lengths between the pretraining and target datasets, we enhance the model's ability to transfer knowledge effectively from pretraining to fine-tuning.
\begin{table}
  \caption{Experimental results in REFinD dev set for model selection}
  \label{tab:model selection result}
  \resizebox{\columnwidth}{!}{%
  \begin{tabular}{lccc}
    \toprule
    Language Model & Micro F1 & Macro F1 & Weighted F1\\
    \midrule
    BERT-base & 84.28 & 68.19 & 84.37 \\
    BERT-large & 85.79 & 70.98 & 85.82 \\
    RoBERTa-base & 85.79 & 69.44 & 85.80\\
    RoBERTa-large & 86.25 & 72.63 & 86.36\\
    ALBERT-base-v2 & 85.35 & 69.64 & 85.22 \\
    ALBERT-large-v2 & 85.62 & 67.41 & 85.59\\
    DeBERTa-base-v3 & 85.58 & 69.59 & 85.64\\
    DeBERTa-large-v3 & \textbf{86.62} & \textbf{72.81} & \textbf{86.65}\\
  \bottomrule
\end{tabular}%
}
\end{table}
\begin{table*}
  \caption{Experimental results on REFinD test set}
  \label{tab:test set result}
  \resizebox{\textwidth}{!}{%
  \begin{tabular}{lcccc}
    \toprule
    Models & Public test Micro F1 & Public test Macro F1 & Public test Weighted F1 & Private test score\\
    \midrule
    Matching the Blanks (BERT-base) \cite{baldini-soares-etal-2019-matching}& 75.19 & 59.36 & 75.67 & - \\
    Matching the Blanks (BERT-large) \cite{baldini-soares-etal-2019-matching}& 76.91 & 62.80 & 77.19 & - \\
    Matching the Blanks (RoBERTa-base) \cite{baldini-soares-etal-2019-matching}& 76.58 & 57.89 & 76.80 & -\\ 
    Matching the Blanks (RoBERTa-large) \cite{baldini-soares-etal-2019-matching}& 78.33 & 61.21 & 78.14 & -\\
    Matching the Blanks (DeBERTa-base) \cite{baldini-soares-etal-2019-matching}& 77.05 & 56.58 & 77.22 & -\\
    Matching the Blanks (DeBERTa-large) \cite{baldini-soares-etal-2019-matching}& 77.60 & 56.34 & 77.54 & - \\
    Matching the Blanks (FinBERT) \cite{baldini-soares-etal-2019-matching, araci2019finbert} & 74.98 & 55.85 & 75.61 & -\\
    Luke-base \cite{yamada-etal-2020-luke}& 67.23 & 42.85 & 67.79 & -\\
    Luke-large \cite{yamada-etal-2020-luke}& 68.28 & 42.55 & 67.63 & -\\
    FinTree (Ours) & \textbf{80.04} & \textbf{66.90} & \textbf{79.60} & -\\
    \midrule
    FinTree ensemble (Ours) & - & - & - & \textbf{72.14}\\
  \bottomrule
\end{tabular}%
}
\end{table*}
\begin{itemize}
\item {\verb|EDGAR Corpus| (About 153M tokens used)}: This corpus includes a collection of financial data obtained from the Electronic Data Gathering, Analysis, and Retrieval (EDGAR) system run by the U.S. Securities and Exchange Commission (SEC). It comprises various financial documents, including annual and quarterly reports, press releases, and disclosure documents.
\item {\verb|SEC Financial Statement and Notes Data Sets| (About 1.2B tokens used)}: These data sets offer a rich aggregation of corporate financial information extracted from XBRL-formatted financial reports. These data sets include detailed numeric and text data from financial statements and accompanying notes.
\end{itemize}

\subsection{Overall structure}
Our work presents FinTree, an advanced model meticulously designed for relation extraction from financial texts. As depicted in Figure \ref{fig_1}, FinTree uses the DeBERTA transformer model and is further enhanced with several methods for achieving exceptional performance.
Notably, instead of leveraging the [CLS] token, a standard for classification tasks, our model strategically employs the [MASK] token's final hidden state. This token is in a predefined query, a text sequence providing key information about the entities we need to predict relation and their respective types. The query is structured as follows: `The relation is \textbf{[MASK]} between \textbf{\{Entity1\}} and \textbf{\{Entity2\}}. The entity of \textbf{\{Entity1\}} is \textbf{\{Entity1\_type\}} and \textbf{\{Entity2\}} is \textbf{\{Entity2\_type\}}.' This unique approach aligns with MLM pretraining strategies, enhancing performance.

After defining the query, we concat it with the original instances. To incorporate the location information of the entities into our model, we add special tokens at the beginning and the end of each entity, a method we refer to as Position Information (PI). The model classifies by applying a linear layer to the last hidden state of the mask token. After training, we adopted Masking Class Post-Processing (MCPP) which masks the class that cannot appear. The REFinD dataset provides entity type information; each relation class includes entity types like `org:date:formed\_on, pers:univ:employee\_of.' If a class's entity type does not match the input data, we mask that class to prevent the model from predicting it. For instance, for entity types `org' and `data,' we mask the relation class `pers:univ:employee\_of.' because the entity type is different. 

In the following sections, we will detail the experimental setup used to evaluate the performance of FinTree and present results that demonstrate its effectiveness in the financial domain.

\section{Experiments}
\subsection{Training Datasets}
We use the REFinD\cite{kaur2023refind} dataset, a unique financial domain relation extraction dataset. The REFinD dataset is generated entirely from financial documents, specifically the 10-X reports of publicly traded companies sourced from the U.S. Securities and Exchange Commission (SEC) website. This dataset contains 28,676 instances with 22 relations amongst eight types of entity pairs. We provide sentences with designated specific entities and their entity types from which a relationship needs to infer. Our task, therefore, is to accurately predict the relationship class from among the specified 22 classes. 
\subsection{Training Details}
We detail our training strategy and parameters below. Our model uses an AdamW \cite{loshchilov2018decoupled} optimizer with the cosine warm-up scheduler and learning rate 1e-5. As REfinD dataset has long sequences, we processed with a maximum length of 1536 tokens. We utilize a batch size of 8 and train our model over five epochs.
We use REfinD train and dev set to train and evaluate public test sets with F1-score metrics in macro, micro, and weighted configurations. And for evaluating our model in private test scores, we modify the seed and ensemble of the same model simply using a hard voting ensemble. 
Furthermore, we use the Adversarial Weight Perturbation (AWP) technique \cite{NEURIPS2020_1ef91c21} during training. Adversarial perturbation to the model weights during training enhances the model robustness and improves generalization by creating a smoother decision boundary around the training instances. We implemented AWP starting from an intermediate stage, specifically from the third epoch onwards.

\section{Results}
\subsection{Performance Results}
We compare FinTree with various language models with the architecture of Matching the Blanks \cite{baldini-soares-etal-2019-matching} and the Luke-base and Luke-large model \cite{yamada-etal-2020-luke}, which has a compelling performance in relation extraction. The evaluation results of the models are in Table~\ref{tab:test set result}. 
As evidenced by the results, our model, FinTree, outperforms competing models, achieving superior scores across the evaluation metrics on the public test. Furthermore, our ensemble model of FinTree demonstrates impressive performance on the private test score. This comparison highlights the strength of FinTree's performance compared with the existing competitive models in relation extraction, such as Matching the Blanks and Luke. Moreover, more adept in the financial domain in comparison with FinBert.

\subsection{Ablation Study}
We present several strategies for training FinTree, including Masking Class Post-Processing (MCPP), Further Pretraining (FP), Position Information (PI), and Adversarial Weight Perturbation (AWP). An ablation study is conducted to investigate the individual contribution of each strategy towards the final model performance, with the results displayed in Table ~\ref{tab:ablation study}.
All the techniques effectively enhance FinTree's performance, as the ablation of any strategy leads to a decrease in performance across all evaluation metrics. Further Pretraining (FP) is the most impactful among these. The absence of FP contributes to the most significant drop in performance, emphasizing its critical role in adapting the model to the financial domain texts.
\begin{table}[t]
\caption{The result of ablation study on REfinD public test set. 
`MCPP' is Masking Class Post-Processing, `FP' is Further Pretrainig, `PI' is Position Information, and `AWP' implies Adversarial Weight Perturbation } 
\centering
\resizebox{\columnwidth}{!}{%
\begin{tabular}{lccc}
\hline
 Models & Micro F1 & Macro F1 & Weighted F1  \\ \hline
FinTree & 80.04 & 66.90 & 79.60 \\
\quad - MCPP & -0.07 & -0.19 & -0.05  \\
\quad - FP & \textbf{-1.70} & \textbf{-4.95} & \textbf{-1.53}  \\
\quad - PI & -0.72 & -3.40 & -0.62  \\
\quad - AWP & -1.33 & -3.38 & -1.08  \\
\quad - (MCPP \& FP \& PI \& AWP) & \textbf{-2.21} & -4.00 & \textbf{-1.70}  \\
\hline
\end{tabular}%
}
\label{tab:ablation study}
\end{table} 
\section{Conclusion}
In this work, we introduced FinTree, a specialized model tailored for relation extraction tasks within the financial domain. Leveraging the transformer encoder-based model DeBERTA and integrating it with the Pattern Exploiting Training (PET) strategy, FinTree aims to address the challenges posed by financial documents' complex and specialized language. 
Our empirical evaluation of the REFinD dataset demonstrated the proficiency of FinTree, as it outperformed standard baseline models. Ablation study of Further Pretraining shows that our model understands financial domain texts well. We found that focusing on the prediction of the [MASK] token effectively aligns the task with the original MLM pretraining paradigm, which results in a more robust model. We also underscored the importance of our novel strategies, which led to a significant enhancement in model performance when integrated. 

\begin{acks}
The author would like to thank Seong-Eun Hong for his meaningful paper review. 
\\
\\
\\
\\
\\
\\
\\
\\
\end{acks}

\bibliographystyle{unsrt}
\bibliography{sample-base}

\begin{thebibliography}{10}

\bibitem{kaur2023refind}
Simerjot Kaur, Charese Smiley, Akshat Gupta, Joy Sain, Dongsheng Wang, Suchetha
  Siddagangappa, Toyin Aguda, and Sameena Shah.
\newblock Refind: Relation extraction financial dataset.
\newblock In {\em In Proceedings of the 46th International ACM SIGIR Conference
  on Research and Development in Information Retrieval(SIGIR ’23), July
  23–27, 2023, Taipei, Taiwan. ACM, New York, NY, USA,}, 2023.

\bibitem{NIPS2017_3f5ee243}
Ashish Vaswani, Noam Shazeer, Niki Parmar, Jakob Uszkoreit, Llion Jones,
  Aidan~N Gomez, \L~ukasz Kaiser, and Illia Polosukhin.
\newblock Attention is all you need.
\newblock In I.~Guyon, U.~Von Luxburg, S.~Bengio, H.~Wallach, R.~Fergus,
  S.~Vishwanathan, and R.~Garnett, editors, {\em Advances in Neural Information
  Processing Systems}, volume~30. Curran Associates, Inc., 2017.

\bibitem{schick-schutze-2021-exploiting}
Timo Schick and Hinrich Sch{\"u}tze.
\newblock Exploiting cloze-questions for few-shot text classification and
  natural language inference.
\newblock In {\em Proceedings of the 16th Conference of the European Chapter of
  the Association for Computational Linguistics: Main Volume}, pages 255--269,
  Online, April 2021. Association for Computational Linguistics.

\bibitem{devlin-etal-2019-bert}
Jacob Devlin, Ming-Wei Chang, Kenton Lee, and Kristina Toutanova.
\newblock {BERT}: Pre-training of deep bidirectional transformers for language
  understanding.
\newblock In {\em Proceedings of the 2019 Conference of the North {A}merican
  Chapter of the Association for Computational Linguistics: Human Language
  Technologies, Volume 1 (Long and Short Papers)}, pages 4171--4186,
  Minneapolis, Minnesota, June 2019. Association for Computational Linguistics.

\bibitem{liu2019roberta}
Yinhan Liu, Myle Ott, Naman Goyal, Jingfei Du, Mandar Joshi, Danqi Chen, Omer
  Levy, Mike Lewis, Luke Zettlemoyer, and Veselin Stoyanov.
\newblock Roberta: A robustly optimized bert pretraining approach.
\newblock {\em arXiv preprint arXiv:1907.11692}, 2019.

\bibitem{Lan2020ALBERT:}
Zhenzhong Lan, Mingda Chen, Sebastian Goodman, Kevin Gimpel, Piyush Sharma, and
  Radu Soricut.
\newblock Albert: A lite bert for self-supervised learning of language
  representations.
\newblock In {\em International Conference on Learning Representations}, 2020.

\bibitem{he2022debertav3}
Pengcheng He, Jianfeng Gao, and Weizhu Chen.
\newblock Debertav3: Improving deberta using electra-style pre-training with
  gradient-disentangled embedding sharing.
\newblock In {\em The Eleventh International Conference on Learning
  Representations}, 2022.

\bibitem{loukas-etal-2021-edgar}
Lefteris Loukas, Manos Fergadiotis, Ion Androutsopoulos, and Prodromos
  Malakasiotis.
\newblock {EDGAR}-{CORPUS}: Billions of tokens make the world go round.
\newblock In {\em Proceedings of the Third Workshop on Economics and Natural
  Language Processing}, pages 13--18, Punta Cana, Dominican Republic, November
  2021. Association for Computational Linguistics.

\bibitem{baldini-soares-etal-2019-matching}
Livio Baldini~Soares, Nicholas FitzGerald, Jeffrey Ling, and Tom Kwiatkowski.
\newblock Matching the blanks: Distributional similarity for relation learning.
\newblock In {\em Proceedings of the 57th Annual Meeting of the Association for
  Computational Linguistics}, pages 2895--2905, Florence, Italy, July 2019.
  Association for Computational Linguistics.

\bibitem{araci2019finbert}
Dogu Araci.
\newblock Finbert: Financial sentiment analysis with pre-trained language
  models.
\newblock {\em arXiv preprint arXiv:1908.10063}, 2019.

\bibitem{yamada-etal-2020-luke}
Ikuya Yamada, Akari Asai, Hiroyuki Shindo, Hideaki Takeda, and Yuji Matsumoto.
\newblock {LUKE}: Deep contextualized entity representations with entity-aware
  self-attention.
\newblock In {\em Proceedings of the 2020 Conference on Empirical Methods in
  Natural Language Processing (EMNLP)}, pages 6442--6454, Online, November
  2020. Association for Computational Linguistics.

\bibitem{loshchilov2018decoupled}
Ilya Loshchilov and Frank Hutter.
\newblock Decoupled weight decay regularization.
\newblock In {\em International Conference on Learning Representations}, 2019.

\bibitem{NEURIPS2020_1ef91c21}
Dongxian Wu, Shu-Tao Xia, and Yisen Wang.
\newblock Adversarial weight perturbation helps robust generalization.
\newblock In H.~Larochelle, M.~Ranzato, R.~Hadsell, M.F. Balcan, and H.~Lin,
  editors, {\em Advances in Neural Information Processing Systems}, volume~33,
  pages 2958--2969. Curran Associates, Inc., 2020.

\end{thebibliography}

\appendix
\end{document}